\documentclass[11pt]{article}

\usepackage[final]{acl}

\usepackage{times}
\usepackage{latexsym}
\usepackage{amsmath}
\usepackage[subtle]{savetrees}

\usepackage[T1]{fontenc}

\usepackage[utf8]{inputenc}

\usepackage{microtype}

\usepackage{inconsolata}

\usepackage{graphicx}

\usepackage{algorithm}
\usepackage{algpseudocode}

\usepackage{booktabs}
\usepackage{multirow}

\usepackage[dvipsnames,table]{xcolor}
\usepackage{colortbl}
\definecolor{berkeleyblue}{HTML}{3B7EA1}
\definecolor{berkeleygold}{HTML}{FDB515}

\newcommand\cy{\cellcolor{berkeleygold!20}}

\title{Anamnesis: An Open-Source Platform for Large-Scale Backstory-Conditioned Survey Simulation}

\author{Song-Ze Yu, Joseph Suh, Serina Chang, David M. Chan  \\
  University of California, Berkeley \\
  \texttt{\{vaclis,josephsuh,serinac,davidchan\}@berkeley.edu}}

\begin{document}
\maketitle
\begin{abstract}
We present \textit{Anamnesis}, an interactive system for demographically controllable survey simulation using large language models. Open-source and designed for \textit{non-technical} users and researchers, Anamnesis enables the prototyping and stress-testing of survey instruments on virtual populations.\footnote{Our system does not replace studies with human participants.} The platform operationalizes the recently introduced Anthology and Alterity frameworks, which use structured narrative backstories to condition model responses, within a unified web interface. It supports open-ended generation, probabilistic demographic resampling, and multimodal (image and audio) surveys. We evaluate the system through two case studies: (1) replicating segments of Pew Research Center’s American Trends Panel (ATP) on political typology and biomedical issues and (2) emulating human preference in the New Yorker Caption Contest. In both cases, \textit{Anamnesis} produces opinion distributions that more closely match real-world survey data than standard persona-prompting baselines, offering a transparent, reproducible, and open-source alternative to proprietary simulation services.
\vspace{0.5 em}

\textbf{Demo:} \url{https://youtu.be/g4l8Y_ilyW0}

\textbf{Platform site:} \url{https://simulate.group}





\end{abstract}

\section{Introduction}

\begin{figure}[t]
    \centering
    \includegraphics[width=\linewidth]{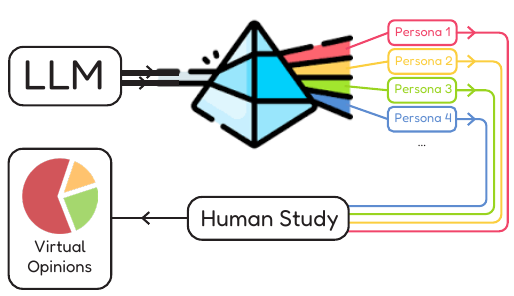}
    \caption{\textit{Anamnesis} is an interactive system for demographically
controllable survey simulation using large language models. It provides a non-technical interface for Anthology, a method which approximates large-scale human studies by conditioning LLMs to representative, consistent, and diverse virtual personas. Together, these systems enable rapid prototyping and stress-testing of survey instruments on diverse virtual populations using multimodal stimuli.}
    \label{fig:teaser}
\end{figure}

Opinion surveys and social polling are foundational tools for understanding human behavior, public policy, and societal trends. However, traditional human-subject research faces mounting challenges, including rising costs, declining response rates, and the logistical difficulty of reaching specific demographic sub-populations. The emergence of Large Language Models (LLMs) as ``virtual personas'' offers an alternative, promising the ability to prototype survey instruments and stress-test social hypotheses at a fraction of the time and cost of traditional methods. For these simulated surveys to be scientifically valid, however, models must move beyond ``average'' aggregate responses and instead demonstrate the ability to faithfully simulate the nuanced, idiosyncratic perspectives of diverse individuals \cite{kang2025deep,moon2024virtual}.

Previous efforts to simulate human populations have primarily relied on ``persona prompting,'' where a model is given a short list of demographic attributes. While functional for basic tasks, this approach often yields stereotypical responses and lacks the psychological depth required for complex opinion elicitation \citep{cheng2023compost}. This limitation has been addressed by the \textit{Anthology} methodology \cite{moon2024virtual} which utilizes rich, open-ended narrative backstories to condition model responses, and the \textit{Alterity} framework \cite{kang2025deep}, which explores ``deep binding'' to ensure LLMs simulate authentic in-group perspectives rather than out-group misperceptions \citep{wang2025large}. Despite these academic advances, the methodologies remain largely confined to siloed Python scripts. Meanwhile, commercial platforms such as \textit{Synthetic Users}, \textit{Expected Parrot}, and \textit{Artificial Societies} \citep{synthetic-users, expected-parrot, yc-artificial-societies} offer similar simulation capabilities but operate as closed-source, proprietary platforms that lack the transparency and reproducibility required for rigorous social science.

In this paper, we present \textbf{Anamnesis}, an open-source, web-based platform designed to democratize access to high-fidelity persona simulation for non-technical users. \textit{Anamnesis} operationalizes the \textit{Anthology} and \textit{Alterity} methodologies within a unified, interactive interface. Unlike previous implementations of these methods, \textit{Anamnesis} is a platform which provides a \textbf{non-technical} survey builder, supports multi-modal inputs (image and audio), and is backed by a range of LLM inference providers. Together, these contributions make state-of-the-art research in persona approximation openly available to a wider range of users. 

We evaluate the \textit{Anamnesis} system through case studies in political opinion elicitation and multimodal preference estimation. Specifically, we replicate segments of the Pew Research Center’s \textit{American Trends Panel} (ATP) \citep{atp}, demonstrating that the platform can elicit opinions that align with real-world human response distributions more accurately than standard prompting baselines \citep{santurkar2023whose, kim2025few}. We also use the platform's multi-modal capabilities to simulate human vision-language preference in the New Yorker Caption Contest. Our results show that \textit{Anamnesis} closely mirrors human sentiment across both language-only and vision-language problems, and can be a valuable tool for researchers prototyping and stress-testing human-study survey instruments.


\begin{figure*}
    \centering
    \includegraphics[width=\linewidth]{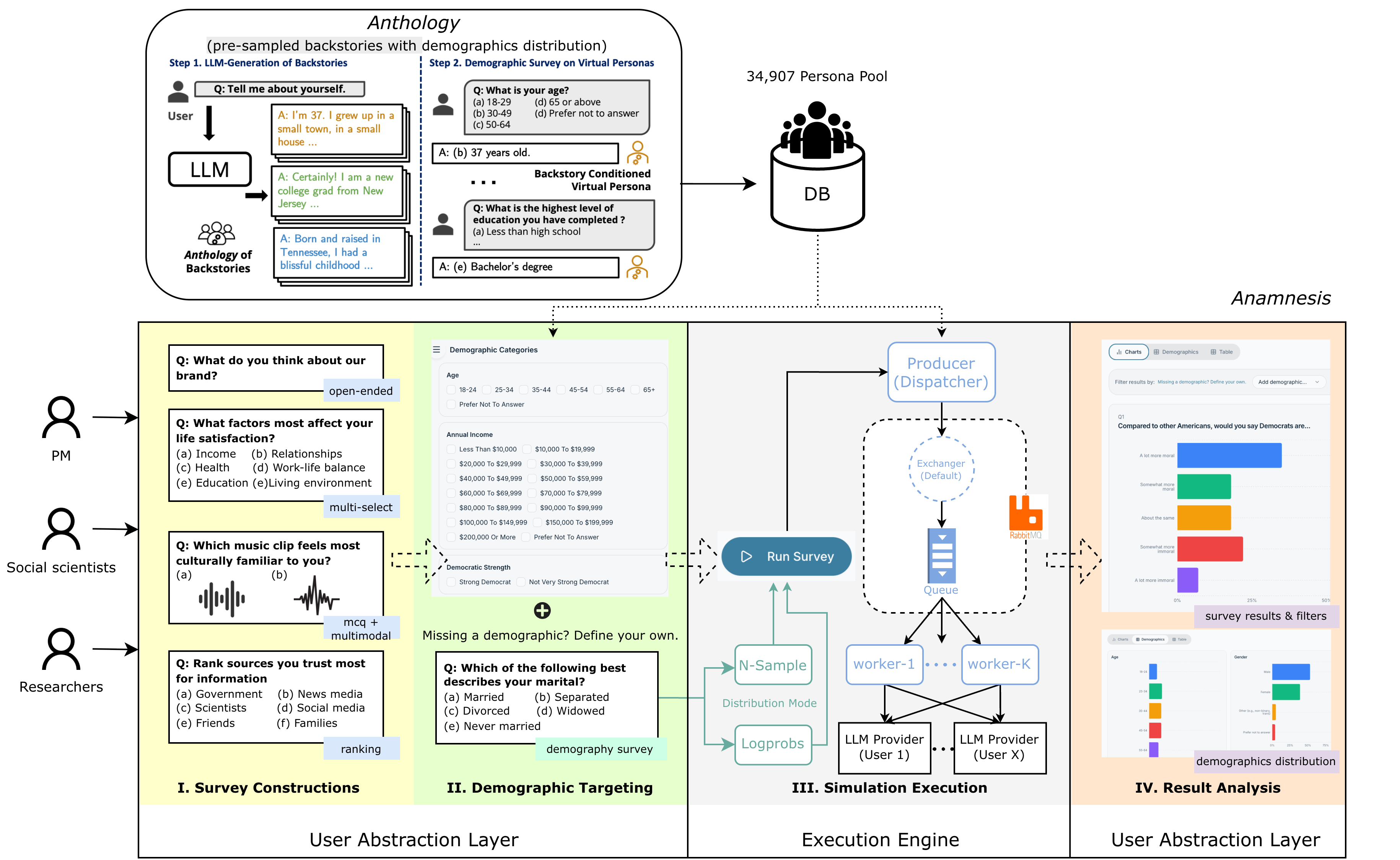}
    \caption{System overview of Anamnesis. Pre-sampled backstories generated via Anthology are stored as a persona pool, each associated with probabilistic demographic distributions. Users construct surveys and specify demographic constraints through an abstraction layer. Survey runs are executed via a dispatcher–queue–worker architecture with sequential context accumulation per persona, enabling scalable and reproducible simulation. Results are aggregated post hoc.}
    \label{fig:architecture}
\end{figure*}

\section{Anthology: Narrative-based Virtual Persona}

Anamnesis is built upon the Anthology framework \citep{moon2024virtual}, which introduces a methodology for simulating diverse human respondents using LLM-conditioned virtual personas.
Rather than relying on short demographic prompts (e.g., "Respond as if you are a 35-year-old Hispanic woman") \citep{santurkar2023whose}, Anthology conditions language models on a rich, open-ended narrative \textit{backstory}-a multi-paragraph life history that captures not just demographic attributes but also formative experiences, values, and worldview.

Backstories are generated via sampling multi-turn life narratives from pretrained base language models \citep{kang2025deep}.
Specifically, the language model is conditioned on interview questions of the American Voices Project \citep{stanford2021american} to complete realistic and diverse open-ended life narratives.
Sampled backstories are labeled by their demographic information which is obtained by querying a multiple-choice demographic question to a language model conditioned on the backstory.
In Anamnesis, we construct a database of pre-sampled backstories indexed by their demographics so that practitioners interested in a subpopulation behavior can easily run a targeted simulation.

To ensure demographic representativeness and a targeted simulation, Anthology pairs backstory generation with a population-matching step.
A practitioner often has a target distribution over demographic dimensions (e.g., age, race, political affiliation) they aim to simulate: to this end, Anthology samples from the entire pool of backstories so that the demographic distribution of the sampled pool matches a target demographic distribution.


Anamnesis operationalizes these methodologies into an end-to-end platform. With a built-in database of indexed backstories, it offers automated backstory generation, demographic balancing based on the target demographics, and response collection with an arbitrary question set to ask a language model conditioned on backstories, all within a single interactive interface.

\section{Anamnesis: Accessible, Open-Source, Anthology Implementation}

\subsection{System Overview}

\textit{Anamnesis} translates the \textit{Anthology} methodology from research prototypes into a deployable open-source survey simulation platform. Rather than requiring researchers to manually generate backstories or write sampling scripts, the platform enables demographic-constrained simulation over a large pool of pre-generated personas through an interactive interface. A typical user workflow consists of four stages:

\begin{itemize}
    \item[1. ] \textbf{Survey Construction:} Users define multi-question survey instruments through a graphical builder. The system supports multiple-choice, multi-select, open-ended, ranking, and multimodal (image and audio) questions.

    \item[2. ] \textbf{Demographic Targeting:} Users specify target audience demographics and sample size, selecting from a pool of 35K pre-sampled backstories with probabilistic demographic distributions (§~\ref{sec:selection}). If a desired demographic dimension is not available, users may create new dimensions through an integrated demographic inference procedure(§~\ref{sec:demog-survey}).

    \item[3. ] \textbf{Simulation Execution:} Users select a language model and answering algorithm(§~\ref{sec:algorithms}). Each backstory completes the survey sequentially, with responses accumulated to maintain consistency (§~\ref{sec:execution}).

    \item[4. ] \textbf{Result Analysis:} Responses are automatically aggregated and visualized. Users may further filter results by demographic attributes post hoc for comparative analysis.
\end{itemize}

\subsection{Execution Architecture}
\label{sec:execution}

As a publicly accessible platform, \textit{Anamnesis} must support concurrent survey runs over a large and growing persona pool. A survey run selects $S$ personas, each of which answers $Q$ questions sequentially, resulting in $O(S \times Q)$ LLM calls before retries and parser calls.
For demographic surveys using repeated sampling ($N$-sample mode; §~\ref{sec:demog-survey}), this yields $O(S \times Q \times N)$ calls.

This execution regime requires (1) per-persona state preservation across questions, 
(2) bounded concurrency under API/vLLM rate limits, and (3) reproducible run-level configuration.

Anamnesis addresses these constraints through a dispatcher–queue–worker architecture (Figure~\ref{fig:architecture}). 
Each survey run is snapshotted at launch time, recording its demographic filters, answering algorithm, model configuration, and concurrency bounds. 
Each selected persona is published to the message queue as one task; 
workers consume tasks asynchronously while still executing questions sequentially within each persona 
with incremental context accumulation. The dispatcher caps the total number of in-flight tasks using the concurrency bounds stored with each run. 

\subsection{Backstory Selection}
\label{sec:selection}

\textit{Anamnesis} enables researchers to simulate surveys over their specified target populations (e.g., “women, aged 18--24” or “voters, aged 25--44, with a college degree, evenly split between Democrat and Republican”) without manual preprocessing. In prior \textit{Anthology} experiments, each backstory was paired with an actual human respondent from a completed real-world survey (e.g., American Trends Panel). Demographic attributes were directly observed. Balancing therefore reduced to deterministic assignment: given known labels and target quotas, one could apply greedy selection or Hungarian matching to choose respondents whose attributes exactly satisfied requested cells.

In \textit{Anamnesis}, this assumption no longer holds. Demographics are not observed labels but inferred probability distributions stored per dimension. For each backstory $b$ and dimension $d$, the system stores:

\[
p_{b,d}(c), \quad c \in \mathcal{C}_d,
\]

where $p_{b,d}(c)$ denotes the inferred probability that $b$ belongs to category $c$. Demographic selection must therefore operate under uncertainty. To accommodate different user scenarios, \textit{Anamnesis} provides two selection algorithms:

\paragraph{Top-K Probability Ranking.}

Designed for scenarios where researchers prioritize selecting personas that most strongly match the target demographic constraints, effectively treating the filtered demographic set as a single group without internal balancing. Given a sample size $S$ and demographic filters, each backstory is scored by its joint compatibility with the filter (multiplying probabilities across dimensions and summing over selected categories when applicable). Backstories are ranked by this score and the top $S$ are selected. 

\paragraph{Balanced Demographic Matching.}

Designed for studies where representation across demographic subgroups must be explicitly enforced (e.g., equal allocation across age $\times$ gender cells). 
First, selected categories are expanded into their cross-product demographic cells $\mathcal{G}$. The total sample size $S$ is divided into slots $K_g$ for each cell $g \in \mathcal{G}$ (uniformly or via user-specified weights). Each slot represents a required demographic target.

Because our pre-sampled backstories include probabilistic demographics, selection becomes an assignment problem: choose backstories such that (i) each slot is filled, (ii) each backstory is selected at most once, and (iii) overall demographic compatibility is maximized. Algorithm~\ref{alg:balanced} summarizes the procedure.




To maintain interactive latency, balanced matching restricts the candidate
space before solving the assignment problem. Without pruning, matching $S$
target slots against the full persona pool $\mathcal{P}$ produces an
$S \times |\mathcal{P}|$ score matrix, which is padded to a square matrix of dimension
$\max(S,|\mathcal{P}|)$ by the Munkres implementation, resulting in
$O(\max(S,|\mathcal{P}|)^3)$ time complexity. Applying this procedure directly
to the full persona pool is impractical for client-side computation.

We therefore retain only the top-$M$ candidate backstories within each demographic cell (default $M=50$) and take their union to form a shared candidate pool $\mathcal{C}$.
Hungarian assignment is then applied over the resulting $S \times |\mathcal{C}|$ matrix, where
$|\mathcal{C}| \ll |\mathcal{P}|$ in typical configurations, making the assignment practical for client-side computation. 
The current interface caps the requested slots at $S\leq50$. Hence, balanced matching is intended for
\textit{small} samples that require \textit{balanced} representation across demographic cells.\footnote{These limits apply only to balanced matching; other selection mode (Top-K) can sweep the whole persona pool.}

\begin{algorithm}[t]
\caption{Balanced demographic matching}
\label{alg:balanced}
\begin{algorithmic}[1]
\Require Persona pool $\mathcal{P}$; filters $\mathcal{F}$; sample size $S$
\State $\mathcal{G} \gets$ cross-product of selected demographic categories
\State Allocate slots $K_g$ for each $g \in \mathcal{G}$ such that $\sum_g K_g = S$
\State Expand slots into target list $\mathcal{T}$ ($|\mathcal{T}| = S$)
\ForAll{$g \in \mathcal{G}$}
    \State Retain top-$M$ backstories by one-hot score for $g$
\EndFor
\State Build score matrix between targets $\mathcal{T}$ and candidate backstories
\State Apply Hungarian assignment to maximize total compatibility
\State \Return matched backstories
\end{algorithmic}
\end{algorithm}

\subsection{Extending the Demographic Space}
\label{sec:demog-survey}

\textit{Anthology} already introduced demographic surveys over backstories, and our persona pool includes pre-populated demographic dimensions derived from that pipeline. \textit{Anamnesis} extends this capability to a user-driven platform feature. Researchers may require attributes not originally annotated (e.g., marital status, political leaning, occupation). Instead of offline scripts, users define a new categorical dimension through the interface, and the system conducts a demographic survey over the persona pool, estimating for each backstory a probability distribution over categories.

\paragraph{Distribution Modes.}

While prior research code relies on token log-probabilities from a self-hosted vLLM backend, many researchers do not operate such infrastructure. We therefore support two interchangeable modes:

\begin{itemize}
    \item \textbf{Logprobs mode (recommended).} When available (e.g., vLLM), a single constrained forward pass yields the full categorical distribution.
    \item \textbf{N-sample mode.} When logprobs are unavailable, the system repeats the question $N$ times and estimates the empirical distribution from sampled responses.
\end{itemize}

Both modes produce the same probabilistic abstraction. Although small $N$ in N-sample mode may introduce sampling variance, the estimate converges as $N$ increases. While this approach incurs higher inference cost than logprobs mode, it closes the practical gap for researchers without access to self-hosted vLLM.

\subsection{Survey Answering Algorithms}
\label{sec:algorithms}

Beyond backstory selection, \textit{Anamnesis} allows users to choose the answering algorithm used during inference, enabling controlled comparisons between simulation strategies.

\paragraph{Anthology (default).} Each backstory is prepended to the first survey
question. After the model responds, the question--answer pair is appended to the
context before the next question is posed, implementing sequential context
accumulation. This mechanism encourages the virtual persona to condition on its prior responses, promoting cross-question belief consistency.

\paragraph{Zero-shot baselines.} Users may alternatively select a baseline mode that conditions only on a short demographic description (e.g., CLAIR-style prompts \citep{chan-etal-2023-clair-evaluating}) rather than the full narrative backstory. Running both modes side-by-side enables direct quantification of the contribution of backstory conditioning, serving as an ablation control.

Responses are parsed through a two-tier pipeline: structured output (guided decoding on vLLM; JSON schema on OpenRouter) is attempted first; if unsuccessful, a lightweight parser LLM extracts the final answer from the raw response.

\subsection{Post-Hoc Demographic Filtering}
\label{sec:postfilter}

For exploratory studies, researchers may execute a survey over the full persona pool without specifying demographic constraints upfront, and subsequently segment results by demographic attributes after the fact. The results dashboard supports interactive filtering and re-aggregation by any demographic dimension stored in the backstory metadata, without requiring re-execution of the survey run.

\section{Case Studies}

We anticipate that practitioners will find diverse applications for Anamnesis, tailoring simulations to their specific needs.
In the following section, we highlight two illustrative use cases and encourage the community to discover further possibilities.

\subsection{Simulating Public Opinion Polls}

To validate that the \textit{Anamnesis} platform replicates the Anthology method, we replicate the core experiment of \citet{moon2024virtual}:
approximating survey response distributions from the Pew Research Center's American Trends Panel (ATP).
We consider three ATP waves covering distinct topics:
Wave 34 (biomedical and food issues), Wave 92 (political typology), and Wave 99 (AI and human enhancement)
(see Appendix \Ref{sec:appendix:ATP} for details).
Survey questions are multiple-choice items asked to all respondents and preserve the original wording and answer options.

Using the Anamnesis survey builder, we construct each ATP wave as a multi-question session.
We configure persona selection in Top-$K$ Probability Ranking mode with no demographic filters. With an empty filter, the platform shuffles the available backstories and returns the requested sample size. A request covering the full pool returns all backstories. Each selected backstory completes the full survey once with sequential context accumulation (§\ref{sec:execution}). Appendix~\ref{sec:appendix:ATP} reports the selected and completed persona counts for each wave.

Following the evaluation protocol of \citet{moon2024virtual}, we map the generated virtual personas to individual ATP respondents using human demographic records and either greedy or maximum-weight assignment. We then compute the average Wasserstein distance (WD) to measure representativeness of the response distribution, and the Frobenius norm between response correlation matrices (Fro.) to measure response consistency.

\autoref{table:atp_reproduce_exp} summarizes the results.
Consistent with the original findings, backstory-conditioned simulation on the Anamnesis platform outperforms demographic list-based baselines across three waves.
Reproducing these three experiments, spanning 20 survey questions and thousands of virtual respondents, the pipeline was configured and executed through the platform's graphical interface.
This highlights the primary utility of Anamnesis:
a social scientist can draft a survey instrument and stress-test it with a diverse virtual population before recruiting human participants.
The platform's interactive result viewer further supports post hoc filtering by demographic subgroup, enabling targeted analysis (e.g., examining whether response distributions diverge across age or race groups) without re-running the simulation.

\begin{table*}[!t]
    \centering
    \scriptsize
    \caption{Simulating American Trends Panel public opinion polls with the Anamnesis platform and two demographic-list prompting methods, BIO and QA \citep{santurkar2023whose}.
    Please refer to \citet{moon2024virtual} for the details of method choices, including persona matching, and the definition of metrics (Wasserstein distance (WD) and Frobenius Norm (Fro.)).
    }
    \vspace{-5pt}
    \label{table:atp_reproduce_exp}
    \resizebox{\textwidth}{!}{
    \begin{tabular}{c|c|c|cc|cc|cc}
    \toprule
    \multirow{2}{*}{Model} & Persona & Persona &
    \multicolumn{2}{c|}{ATP Wave 34} & \multicolumn{2}{c|}{ATP Wave 92} & \multicolumn{2}{c}{ATP Wave 99}
    \\
    & Conditioning & Matching &
    WD ($\downarrow$) & Fro. ($\downarrow$) &
    WD ($\downarrow$) & Fro. ($\downarrow$) &
    WD ($\downarrow$) & Fro. ($\downarrow$)
    \\
    \midrule
    \multirow{5}{*}{LLaMA-3.1-8B}
    & BIO & n/a & 0.258 & 1.556 & 0.346 & 2.078 & 0.277 & 1.229
    \\
    & QA & n/a & 0.235 & 1.481 & 0.392 & 1.719 & 0.180 & 1.475
    \\
    & \cy & \cy max weight & \cy 0.160 & \cy 0.837 & \cy 0.251 & \cy 1.603 & \cy 0.148 & \cy 1.026
    \\
    & \multirow{-2}{*}{\cy Anamnesis} &\cy greedy & \cy 0.147 & \cy 0.964 & \cy 0.218 & \cy 1.414 & \cy 0.139 & \cy 1.352
    \\
    \midrule
    \multicolumn{3}{c|}{Human} &  0.057 & 0.418 & 0.091 & 0.411 & 0.081 & 0.327
    \\
    \bottomrule
    \end{tabular}
    }
\end{table*}

\subsection{Multimodal Alignment}
\label{sec:case-study}

Prior experiment only focused on text-based surveys. To verify that backstory-based simulation remains meaningful under multimodal inputs, we evaluated alignment against real human preference data.

We therefore conduct a case study on the New Yorker Caption Contest benchmark \citep{hessel2022androids, newyorkernextmldataset}, a multimodal task in which cartoon images are paired with caption candidates and ground-truth labels are derived from large-scale crowd voting. The dataset provides a simple but controlled test of whether persona-conditioned virtual populations exhibit measurable correlation with collective human judgments.

\paragraph{Method.}
We evaluated 49 contests with randomized caption order. For each, Gemini~2.5~Flash (temperature 1.0) makes 20 choices under two answering algorithms: (i) Anthology (backstory-conditioned simulation) (ii) Zero-shot demographic baseline. We report majority-vote accuracy with Wilson intervals and an exact McNemar test. To retain within-item information, we also compare the vote share assigned to the human winner using a paired bootstrap interval and exact sign-flip test.

\paragraph{Results.}

Anthology achieves a majority-vote accuracy of 59.2\% (95\% CI: 45.2--71.8\%), compared with 51.0\% (95\% CI: 37.5--64.4\%) for the Zero-shot baseline. Narrative conditioning also increases the mean vote share assigned to the human-preferred caption from 52.0\% to 59.8\%, a paired improvement of 7.8 percentage points (95\% CI: 3.2--12.8; $p=0.0024$). These findings suggest that Anthology shifts model preferences toward the human-preferred caption overall.

\section{Related Work}
\label{sec:related_work}

\paragraph{LLM Persona Conditioning.}
A growing body of work explores conditioning LLMs to simulate human perspectives.
Early approaches supply language models with short demographic attribute lists, e.g., question-answer pairs about demographic indicators, and measure alignment with human survey responses \citep{santurkar2023whose, hwang2023aligning, li2025llm}.
While effective as baselines, these methods tend to produce stereotypical or flattened outputs that fail to capture within-group variation \cite{cheng2023compost, wang2025large}.
Anthology \cite{moon2024virtual} advances this line by conditioning on rich, LLM-generated narratives rather than attribute lists, demonstrating improved consistency on survey benchmarks; Alterity \cite{kang2025deep} further demonstrates the efficacy via reproducing in-group, out-group and meta-perception study results.

\paragraph{Comparison to Existing Survey Platforms.}

A growing ecosystem of commercial and open-source platforms offers LLM-based survey simulation.
\emph{Synthetic Users} \cite{synthetic-users} generates AI personas for interviews and surveys using a multi-agent framework with optional retrieval-augmented generation to incorporate proprietary data.
However, personas are defined by short attribute profiles rather than rich narratives, and the methodology is entirely closed-source.
\emph{Artificial Societies} \cite{yc-artificial-societies} focuses on simulating a network-level social dynamics,
such as content virality
and collective decision-making, rather than structured opinion surveys, and likewise does not publish its conditioning methodology to simulate virtual personas.



On the open-source side, Expected Parrot's EDSL \cite{expected-parrot} provides a Python domain-specific language and command-line workflow for constructing AI agents and surveys. 
Although EDSL can condition personas through trait dictionaries rendered by customizable Jinja2 templates, a long narrative backstory must be supplied by the user as a trait. Moreover, as a code-oriented toolkit, it remains less accessible to researchers without programming experience, as it requires explicit code-based specification of scenarios, user/agent models, tools, policies, and evaluation hooks.

\citet{park2024generative} demonstrate that two-hour qualitative interviews with real individuals can produce generative agents that replicate survey responses with high fidelity, though this approach requires costly human data collection that limits scalability.
Anamnesis is, to our knowledge, the first open-source, GUI-based platform that combines narrative backstory conditioning, probabilistic demographic matching, sequential context accumulation, and multimodal survey support together with a pre-generated pool of 34,907 narrative personas in a single deployable system accessible to researchers without programming expertise.



\section{Limitations and Ethics Statements}
\label{sec:app:limtiations}

While \textit{Anamnesis} provides a robust platform for persona-based survey simulation, several limitations inherent to the methodology and the underlying technology must be acknowledged. First, the quality of any simulation is fundamentally bounded by the diversity of the backstory pool. As identified in the development of the \textit{Anthology} framework \cite{moon2024virtual}, LLM-generated backstories can exhibit skewed demographic distributions that reflect the inherent biases of their training data, not exactly the true census-representative population. This leads to the risk of ``shallow binding,'' where the model reflects an out-group's stereotypical perception of a demographic rather than the group's actual internal logic. Although \textit{Anamnesis} implements methodologies from the \textit{Alterity} framework \cite{kang2025deep} to deepen this binding through multi-turn interview transcripts, researchers should remain critical of results on sensitive social topics where models may still default to caricatured personas. Moreover, because all backstories and simulations are conducted in English, linguistic and cultural variation is necessarily compressed into English-language reasoning patterns, potentially limiting the cross-cultural validity of represented personas.


Additionally, while the platform enables multi-modal conditioning, current multi-modal LLMs (MLLMs) may lack the perceptual nuance of human subjects, potentially ignoring subtle visual or auditory cues. Finally, virtual personas are temporally static; they do not evolve in response to real-world current events unless their backstories are explicitly updated, which limits the platform's utility for longitudinal tracking of rapidly shifting public opinion.

Last but not least, \textit{Anamnesis} is intended for survey prototyping and stress-testing. Its outputs are model-generated predictions, not observations of human attitudes. We don't claim that our system can replace studies with human participants. The demographic fields are also inferred by a language model and may contain errors or stereotypical associations. Researchers should still validate substantive and trustworthy findings with human participants.

\bibliography{custom}

\section*{Acknowledgements}

As part of their affiliation with UC Berkeley, the authors were supported in part by the U.S. Department of War, the Berkeley Artificial Intelligence Research (BAIR) Industrial Alliance program, and Veridian Technologies. This material is based upon work supported by the Defense Advanced Research Projects Agency and the Air Force Research Laboratory under contract FA8650-23-C-7316. We thank Yutong Bai for her comments and thoughts on an earlier version of this work. Any opinions, findings, conclusions, or recommendations expressed in this material are those of the authors and do not necessarily reflect the views of AFRL or DARPA.
\appendix
\counterwithin{figure}{section}
\counterwithin{table}{section}

\section*{Appendix}

\begin{itemize}
    \item \autoref{sec:appendix:pool} reports the probabilistic demographic composition of the persona pool.
    \item \autoref{sec:appendix:ATP} discusses some additional details of the American Trends Panel.
    \item \autoref{sec:appendix:multimodal} discusses some additional details of the New Yorker Caption Contest.
    \item \autoref{sec:appendix:ai-disclosure} provides the AI usage disclosure.
\end{itemize}
\vfill
\pagebreak

\section{Persona Pool Demographics}
\label{sec:appendix:pool}
The public persona pool contains 34,907 Alterity backstories. A backstory is not assigned one fixed demographic identity. For each demographic dimension $d$, it stores the inferred distribution $p_{b,d}(c)$ defined in §~\ref{sec:selection}. Operationally, each backstory contributes fractional probability mass to several demographic subpopulations, allowing one persona to represent a small, demographically mixed group of people.

\begin{figure}[H]
    \centering
    \includegraphics[width=\linewidth]{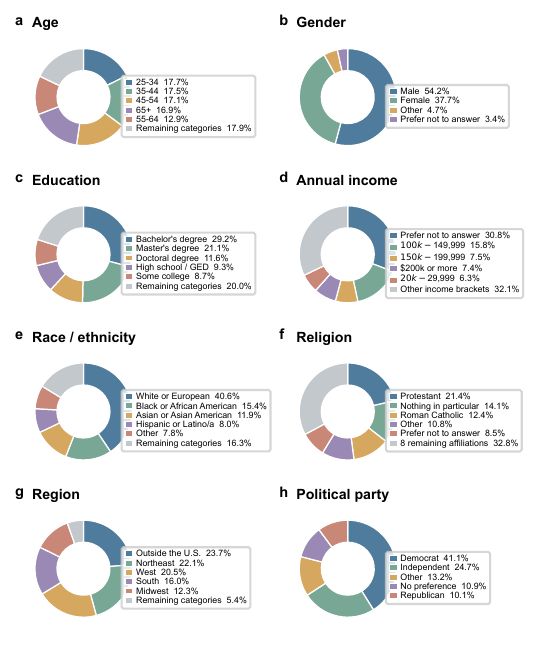}
    \caption{\textbf{Probabilistic demographic composition of the persona pool.}
    The figure aggregates the probability mass assigned to each demographic category across 34,907 backstories; each backstory may contribute to multiple categories. The income remainder (32.1\%) combines nine brackets: below \$10,000, \$10,000--\$19,999, and seven \$10,000-wide brackets covering \$30,000--\$99,999. The religion remainder (32.8\%) combines Jewish, atheist, Muslim, Mormon/LDS, agnostic, Buddhist, Hindu, and Orthodox affiliations.}
    \label{fig:persona-pool-demographics}
\end{figure}

Figure~\ref{fig:persona-pool-demographics} reports the pool-level share for category $c$ as $N_d^{-1}\sum_{b\in\mathcal{B}_d}p_{b,d}(c)$, where $\mathcal{B}_d$ is the set of backstories annotated for dimension $d$. The figure therefore aggregates probability mass, not top-choice labels or counts of uniquely classified personas. These attributes are inferred from the backstories by a language model; they are not self-reported demographics or estimates of a real population.

\vfill
\pagebreak

\section{American Trends Panel}
\label{sec:appendix:ATP}
 \label{appendix:ATP_W99}

The American Trends Panel (ATP), administered by the Pew Research Center \citep{atp}, is a nationally representative survey panel comprising U.S. adults.
The panel covers a broad range of subjects, from politics and religion to internet use and online dating, among others.
Our analysis draws on selected questions from three survey waves, focusing on items that were posed to all human participants.
Notably, some questions in the original ATP surveys use Likert-scale response options whose ordering (e.g., ranging from positive to negative, or vice versa) was randomized across respondents.
To mirror this design, we similarly randomize the sequence of these options when constructing prompts for LLMs.

\begin{figure*}
    \centering
    \includegraphics[width=0.95\linewidth]{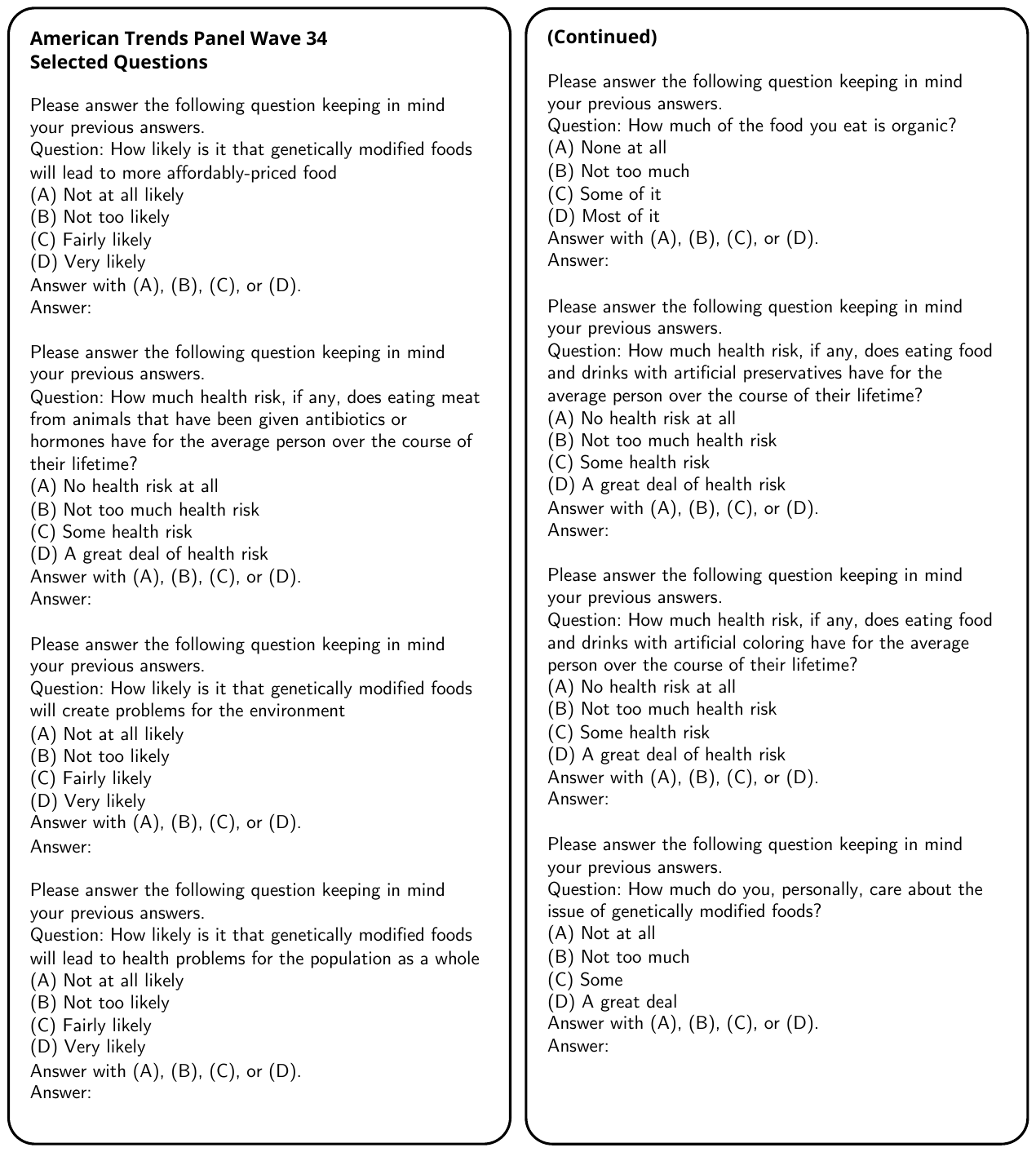}
    \vspace{-35pt}
    \caption{Eight questions sampled from ATP Wave 34.
    The prompts ``Please answer the following question keeping in mind your previous answers'' are included before asking each survey question, which are found to enhance the consistency of response from \citet{moon2024virtual}.}
    \label{fig:atp_34_prompts}
\end{figure*}

Figure~\ref{fig:atp_34_prompts} provides an example of the survey prompts and sequential-context instruction used in these experiments.
ATP Wave 34 is conducted from April 23, 2018 to May 6, 2018 with a focus on biomedical and food issues. The number of total respondents is 2,537. An example is provided in Figure~\Ref{fig:atp_34_prompts}. ATP Wave 92 is conducted from July 8, 2021 to July 21, 2021 with a focus on political typology and 10,916 respondents. American Trends Panel Wave 99 is conducted from November 1, 2021 to November 7, 2021 with a focus on artificial intelligence and human enhancement. The number of total respondents is 10,260. 

\begin{table}[h]
\centering
\scriptsize
\begin{tabular}{lrrrr}
\toprule
Wave & Questions & Human $n$ & Selected & Completed \\
\midrule
34 & 8 & 2,537 & 5,000 & 5,000 \\
92 & 7 & 10,916 & 34,907 & 34,905 \\
99 & 6 & 10,260 & 34,907 & 34,905 \\
\bottomrule
\end{tabular}
\caption{Selected and completed persona counts refer for each wave.}
\label{tab:atp-run-scale}
\end{table}


\section{New Yorker Caption Contest}
\label{sec:appendix:multimodal}

\begin{figure}[H]
  \centering
   \includegraphics[width=\linewidth]{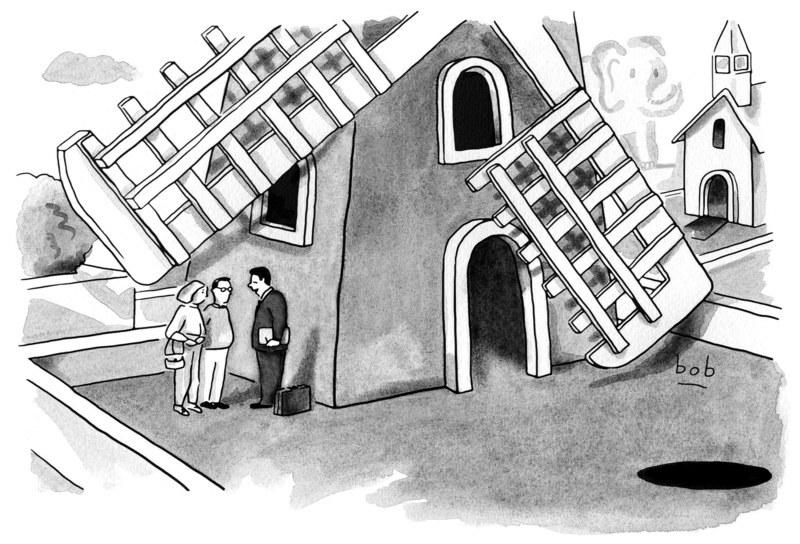}
  \caption{\textbf{Image-based caption ranking example.}
  Given the cartoon image above, the model must select the funnier caption among the candidates:
  (A) ``It comes with sub-par schools but a world-class trauma center.''
  (B) ``If we time it right, I can get you in this house today.'' (GT: B)}
  \label{fig:nycc_example}
\end{figure}

The New Yorker Caption Contest Benchmarks dataset \cite{hessel2022androids} is a large-scale multimodal benchmark designed to evaluate computational ``humor understanding'' using cartoons from The New Yorker Caption Contest. We evaluate our method on the ``Quality Ranking'' task, which requires methods to choose the funnier caption between alternatives. Each instance includes the original cartoon image, two captions, and gold labels for which caption won the contest. The dataset supports image-based and text-based settings; we use the image-based version for our experiments. An example is given in \autoref{fig:nycc_example}.

\section{AI Usage Disclosure}
\label{sec:appendix:ai-disclosure}
The authors used AI-based tools to assist with code generation, editing, and writing during the preparation of this paper. Specifically, AI assistance was used to help draft and revise portions of the manuscript for clarity, grammar, and organization, and to support the development, debugging, and refinement of code used in the research workflow. All AI-generated or AI-assisted content, code, analyses, and interpretations were reviewed, verified, and, where necessary, modified by the authors. The authors take full responsibility for the accuracy, integrity, originality, and final content of the paper, including any code or text developed with AI assistance.

%






\end{document}